\documentclass[letterpaper, 10 pt, conference]{ieeeconf}

\IEEEoverridecommandlockouts
\usepackage[pdftex]{graphicx}
\usepackage[tight]{subfigure}
\usepackage{algorithmicx}
\usepackage{algpseudocode}
\usepackage{algorithm}
\usepackage{amssymb,amsmath}
\usepackage{tikz}
\usepackage{pgfplots}
\usepackage{environ}
\tikzset{
  treenode/.style = {shape=rectangle, rounded corners,
                     draw, align=center,
                     top color=white, bottom color=blue!20},
  root/.style     = {treenode, font=\Large, bottom color=red!30},
  env/.style      = {treenode, font=\ttfamily\normalsize},
  dummy/.style    = {circle,draw}
}
\makeatletter
\newsavebox{\measure@tikzpicture}
\NewEnviron{scaletikzpicturetowidth}[1]{%
  \def\tikz@width{#1}%
  \def\tikzscale{1}\begin{lrbox}{\measure@tikzpicture}%
  \BODY
  \end{lrbox}%
  \pgfmathparse{#1/\wd\measure@tikzpicture}%
  \edef\tikzscale{\pgfmathresult}%
  \BODY
}
\makeatother

\PassOptionsToPackage{hidelinks}{hyperref}
\AtBeginDocument{}

\title{\LARGE \bf Physics-Based Damage-Aware Manipulation Strategy Planning\\Using Scene Dynamics Anticipation}

\author{Tobias Fromm and Andreas Birk 
\thanks{The authors are with the Robotics Group, Computer Science \& Electrical Engineering, Jacobs University Bremen, Germany; \emph{t.fromm@jacobs-university.de}. The research leading to the presented results has received funding from the European Union's Seventh Framework and Horizon 2020 programs within the projects ``Cognitive Robot for
Automation Logistics Processes'' (RobLog) and ``Effective Dexterous ROV Operations in Presence of Communication Latencies'' (DexROV).
}
}
\begin{document}
\maketitle
\begin{abstract}
We present a damage-aware planning approach which determines the best sequence to manipulate a number of objects in a scene.
This works on task-planning level, abstracts from motion planning and anticipates the dynamics of the scene using a physics simulation. Instead of avoiding interaction with the environment, we take unintended motion of other objects into account and plan manipulation sequences which minimize the potential damage.
Our method can also be used as a validation measure to judge planned motions for their feasibility in terms of damage avoidance.
We evaluate our approach on one industrial scenario (autonomous container unloading) and one retail scenario (shelf replenishment).
\end{abstract}

\section{Introduction}
Since no robot is perfect, just like the humans building them, autonomous manipulation can be destructive if fragile goods have to be handled. Many compliant grippers, sophisticated perception systems and manipulation routines have been developed recently, mostly relying on consequent obstacle avoidance. This works to a certain degree, but why not take alteration of the environment into account for planning instead of bluntly avoiding interaction with it?

We present a manipulation strategy planning approach which validates and selects the best sequence to remove or unload a number of objects from a scene while taking into account possible damage-prone movements of any other objects. The idea relies on anticipating the dynamics of the scene using a physics simulation which contains a copy of the physical scene, including the robot, the environment and a number of objects to manipulate.

The possibilities are obvious: One the one hand, motion planning is facilitated because of a less constrained search space in contrast to regarding all other objects as obstacles. In some scenes, it may be hard or even impossible to perform manipulation if no collisions are permitted. On the other hand, we want to avoid damage to the possibly heavy or fragile goods as well as the robot itself which may occur if an object is shifted or dropped unintendedly.

Hence, the goal of our method is to optimize a robot's autonomous behavior according to the anticipated dynamics of the real-life scene. This works without imposing explicit spatial or logical dependencies between objects, thus the approach is not tied to a particular application domain.

\begin{figure}[tbp]
  \centering
  \subfigure[Logistics (container unloading) \cite{Stoyanov2016}]{
    \label{fig:scenario_roblog}
    \includegraphics[width=0.78\linewidth]{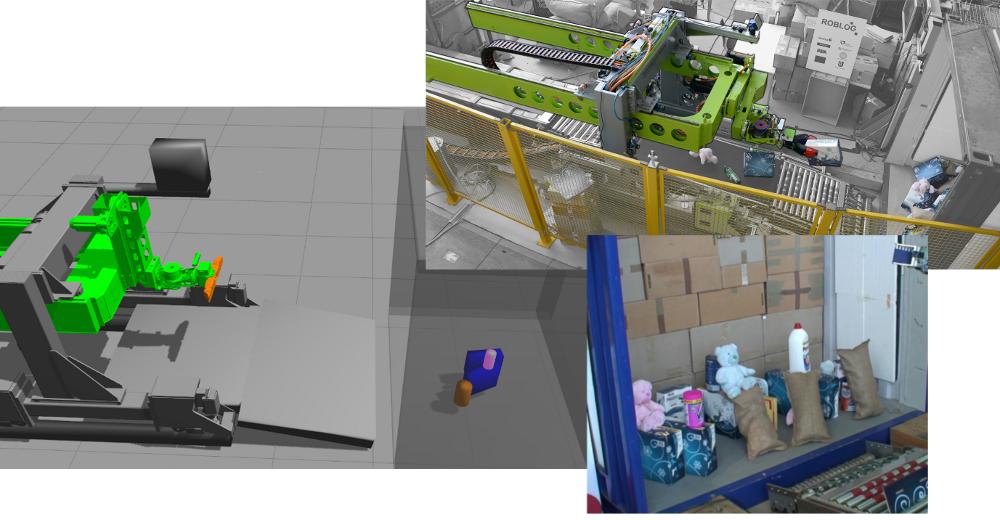} 
  }
  \subfigure[Supermarket (shelf replenishment) \cite{Winkler2016}]{
    \label{fig:scenario_shopping}
    \includegraphics[width=0.78\linewidth]{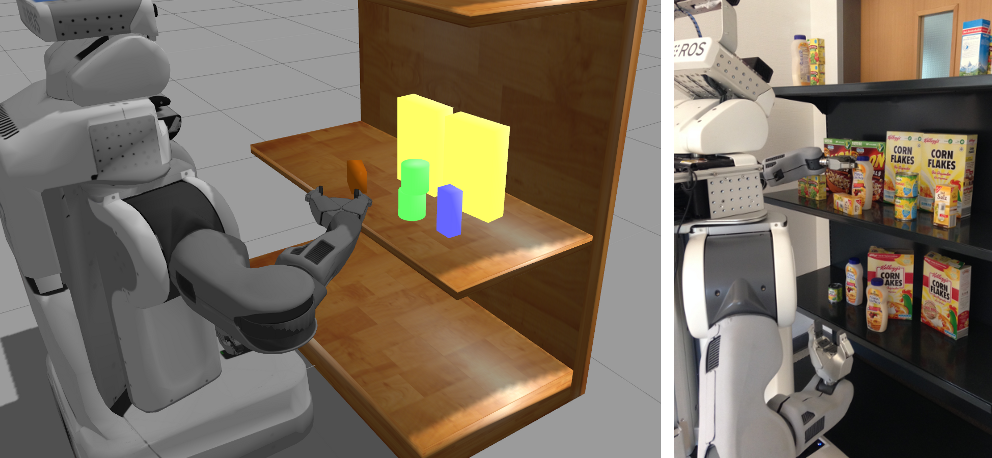} 
  }
  \subfigure[Underwater (maintenance, archaeology and biology) \cite{Gancet2016}]{
    \label{fig:scenario_underwater}
    \includegraphics[width=0.38\linewidth]{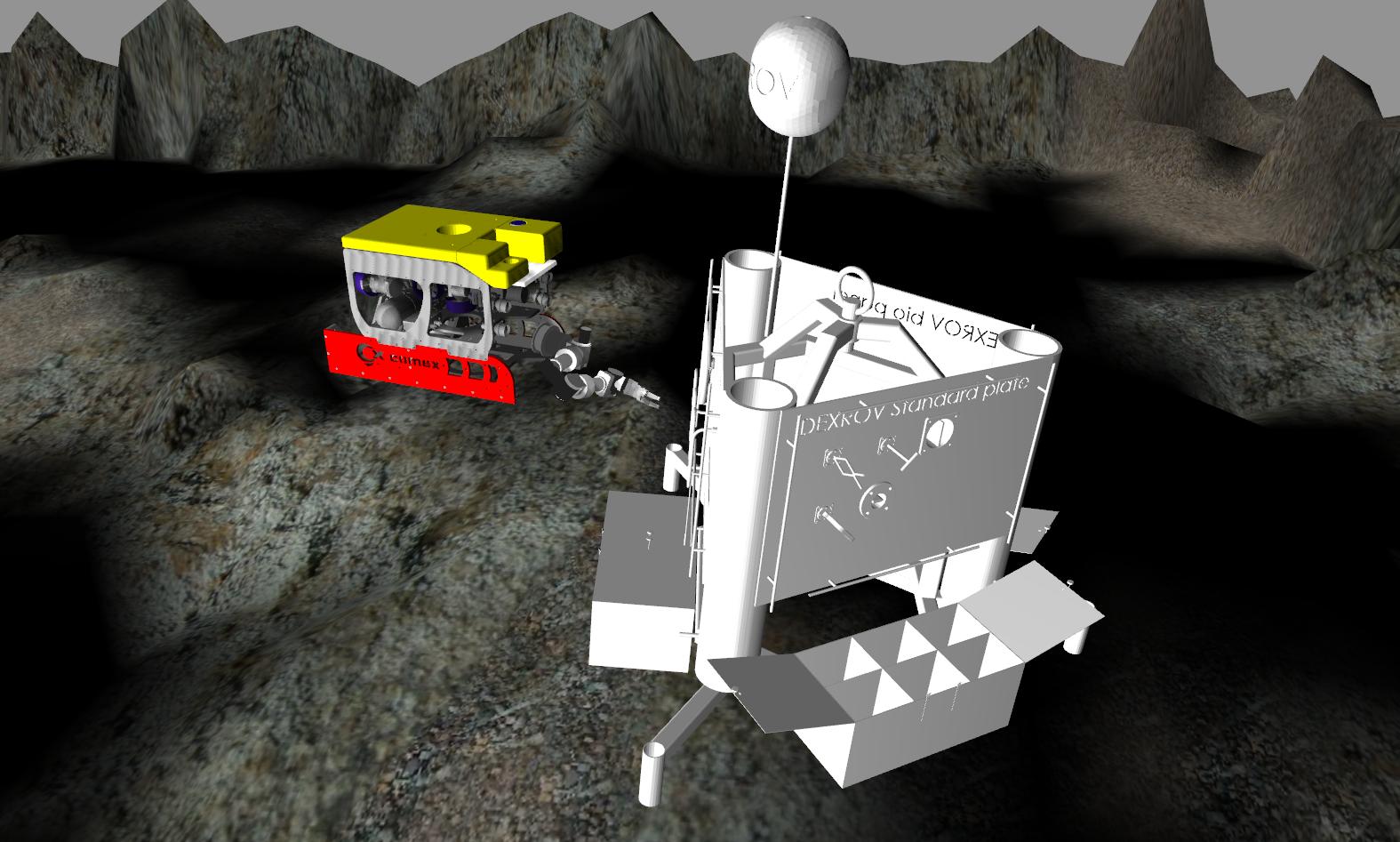} 
    \includegraphics[width=0.38\linewidth]{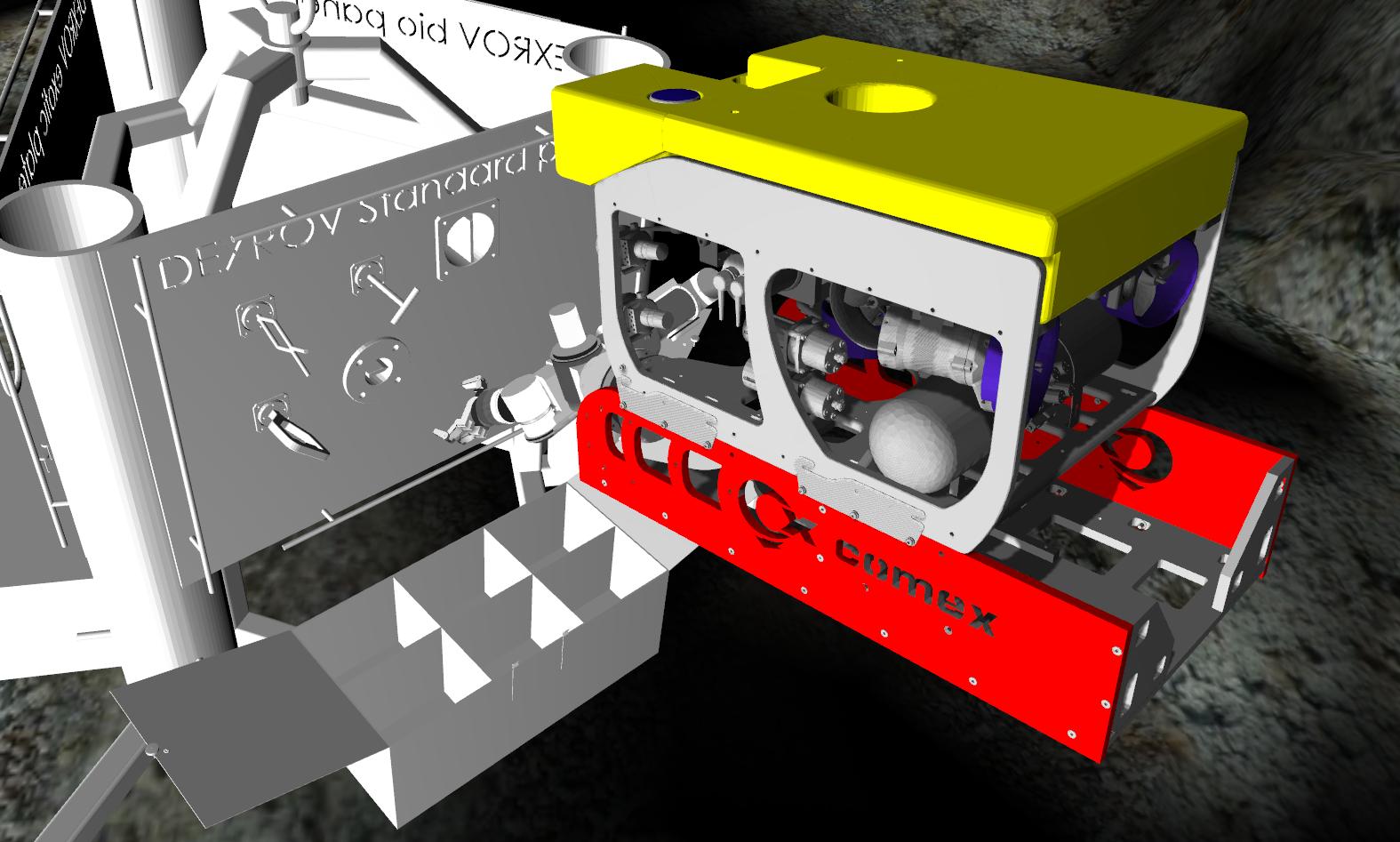}
  }
  \caption{Typical application scenarios for autonomous manipulation}
  \label{fig:scenarios}
\end{figure}

We evaluate our approach on two everyday scenarios shown in Figure~\ref{fig:scenarios}: logistics, in the context of the EU project ``Cognitive Robot for Automation of Logistics Processes'' (RobLog) \cite{Stoyanov2016} where containers are unloaded autonomously, and in a supermarket for shelf replenishment \cite{Winkler2016}.

Since our method is generally applicable for robotic manipulation tasks, underwater environments like used within the EU project ``Effective Dexterous ROV Operations in Presence of Communication Latencies'' (DexROV) \cite{Gancet2016} may serve as another application example.

\textbf{The contribution of our work is:}
\begin{itemize}
    \item a method to plan strategies for the manipulation of a sequence of objects
    \item minimizing unintended object movement while dynamically considering the physics of all objects
    \item including an industrial and a retail application scenario which show that our approach can be utilized in arbitrary usage domains.
\end{itemize}

\section{Foundations and Related Work}\label{related_work}

\subsection{Physics-Based Motion Planning}
Motion planning based on physics simulation has been investigated in several publications, amongst these, Zickler and Veloso's work \cite{Zickler2009}. Instead of conventionally planning motions for a robot's physical degrees of freedom, they group low-level motions and use them as a single point in action selection sampling space.
As for the environment representation, Zickler and Veloso state that it is necessary to define domain-appropriate distance functions between states of the scene in order to be able to compare them. For their usage domain of robot soccer and minigolf, the Euclidean distance serves as a sufficient measure in most cases.

Weitnauer et al.\ \cite{Weitnauer2010} present an approach which works in a similar way, though only on plain objects and using only the Euclidean distance as a distance function.

Another usage example of physics-based planning can be found in \cite{Dogar2012} and \cite{Kitaev2015} where the authors present methods of planning grasps through clutter, taking shifting objects into account and explicitly manipulating them. This is rather similar to our approach, though not with the intention to find an efficient manipulation order, but to clear manipulation paths from obstructing objects.

Before using one of these approaches, the object to manipulate would have to be selected by some algorithm or a human operator. On the contrary, our method automates exactly this scenario where not only physics-based manipulation planning takes place, but also autonomous object selection to plan manipulation sequences on a high level.

In addition, in the scenarios intended to be tackled with our approach, intentionally pushing objects aside with the gripper poses an additional risk of damaging the goods and even the gripper. For this reason, we want to limit the movement of any other than the object to manipulate as much as possible. Our approach takes care of this by penalizing any movement of a \emph{passive object}, i.e. an object which is not subject to intended manipulation at the moment. Therefore, intentional as well as unintentional damage-prone manipulation of passive objects is avoided.

\subsection{Manipulation Strategy Planning}
In contrast to the methods described so far, our approach does not aim on planning \emph{motions}, but high-level \emph{behaviors} with the help of a physics simulation. This means that, instead of giving an explicit trajectory, our method will a) \emph{validate} whether an ongoing manipulation is likely to be successful or b) \emph{plan} a sequence of manipulation actions to perform in the desired scenario. Therefore, our possible application scenarios are separated into \emph{Manipulation Outcome Prediction} and \emph{Manipulation Sequence Planning} which will be introduced in the following.

\paragraph{Manipulation Outcome Prediction}
Covered by this term, several methods have shown different ideas and scenarios. Unfortunately, most of the existing work focuses on predicting the outcome of motions instead of behaviors, like Pastor et al.'s work \cite{Pastor2011} where robot parameters like joint positions and velocities are continuously monitored and verified whether they lie within a learned envelope. Haidu et al.~\cite{Haidu2015} use a similar technique, also to learn motoric skills.

In contrast to these works, Rockel et al.~\cite{Rockel2015} present an interesting idea of predicting task success on the basis of motion of the manipulated objects. Concretely, their application includes a robot which maximizes its speed moving around in a scene, but without losing the balance of an object which sits on its base. This works on the basis of semantic predicates, but with hand-crafted features which describe if the object shakes or topples and which cannot easily be employed on a different scenario.

Akhtar et al.~\cite{Akhtar2013} also use semantic predicates to reason about success of a manipulation action. They present a Machine Learning algorithm which is able to predict whether an object released on top of another will behave as expected. This detection of external faults can predict the behavior of a concrete action based on a simulated training set, but again uses hand-crafted predicates which only cover this particular action. Our approach, on the contrary, aims on covering implicit manipulation of passive objects as well, independent of which exact manipulation is performed on the scene.

\paragraph{Manipulation Sequence Planning}
In order to manipulate objects which are placed currently unreachable behind others, Stilman et al.\ \cite{Stilman2007} use a sampling-based planner to move away the blocking objects.

Okada et al.\ \cite{Okada2004} present a similar simple, but effective example for strategy planning for a humanoid robot where obstacles are considered as planning goals and subsequently moved away from the desired walking path. Their goal is to enable robots to implicitly perform necessary manipulation actions, even if these actions were not part of the original task plan.
This is similar to our recent work evaluated in a different scenario \cite{Winkler2016}, but different to what we present in this work. Here we use a dynamics simulation to evaluate the feasibility of a manipulation action, taking unintended side effects into account in addition to static spatial knowledge.

\section{Prerequisites}\label{overview}
\setcounter{paragraph}{0}

Simulating the behavior of robots and objects requires both high accuracy and some abstractions since not every detail in robot and object properties can easily be modeled.
Since our approach is intended to work in complex scenarios based on perception systems for object recognition and real robots for motion execution, we need to be sure that these work error-free and take care of the respective noise where it occurs. Therefore, we assume object models and environment to be given as well as the state of the robot.

\subsection{Simulation Scene Composition}
In order to build the scene configuration used in the physics simulation, we first recognize and localize all objects in the scene using our perception system \cite{Stoyanov2016} \cite{Vaskevicius2014}. This system includes a pipeline of several segmentation and filtering steps before it executes diverse recognition modules, including a feature-based textured object recognition module \cite{Vaskevicius2012} and a graph-based shape model recognition module \cite{Mueller2014}.

\subsection{Motion Planning, Grasp Planning and Execution}
After all objects have been perceived, the respective grasps need to be planned, evaluated and finally executed by the robot. The way of determining grasping configurations strongly depends on the objects and gripper used in the scenario since different object sizes and structure demand for different gripper sizes and amounts of dexterity. Since the focus in this work does not lie on grasp and motion planning and our method is supposed to be re-used in different scenarios, we use simple grasping configurations around the principal object axes.

The physics simulation uses a kinematics and dynamics model of the robot which has been developed together with the respective simulated controllers. For the supermarket scenario, we use the standard PR2 simulation model. The eventual generation of the motion trajectories can be performed, for example, by using \emph{MoveIt!}.

\section{Manipulation Cost Functions}\label{cost_evaluation}
\setcounter{paragraph}{0}
\subsection{Motivation}
The usage of cost functions to find the most suitable and reasonable solution to a planning problem has been common practice for a long time (e.g.\ in \cite{Okada2004}, \cite{Goldberg1990}) to be able to evaluate the effects of a particular action. In order to apply this method to a specific planning problem, domain knowledge has to come into play which accounts for the optimization target of the respective problem.

Since our approach aims on reducing unintended motions of \emph{passive objects}, i.e. objects which are not subject to intended manipulation at the moment, we need to consider cost functions which cover spatial modifications in the scene.

\subsection{Terminology}
For all of the following definitions, we use $\mathcal{O}$ as the set of objects present in the scene, $\alpha \in \mathcal{O}$ as the \emph{active object} (which will be manipulated), $\Phi = \mathcal{O} \setminus \alpha$ as the set of \emph{passive objects} (which will not be manipulated) and $\phi \in \Phi$ as one member of this set.

\subsection{Pose-Based Cost Functions}
The following two simple cost functions are based on object poses and allow for a quick validation without a lot of processing, but may not be predictive-efficient enough.

\paragraph{Maximum Pose Shift $c_{p}$}\label{cost_evaluation:max_ps}
describes the maximum Euclidean distance between object poses:
\begin{equation}\label{eq:max_ps}
	c_{p} = \max_{\phi \in \Phi}{\lVert \mathbf{p}_{t+1}(\phi) - \mathbf{p}_{t}(\phi)\rVert}
\end{equation}
where $\mathbf{p}_{t}(\phi) \in \mathbb{R}^{3}$ is the position of passive object $\phi$ before (at time $t$) and $\mathbf{p}_{t+1}(\phi)$ after running the simulation step (at time $t+1$). This is simple and easy to compute, but smoothes out non-straight motion paths and thus does not describe changes in an object's direction of movement.

\paragraph{Maximum Path Length $c_{l}$}\label{cost_evaluation:max_pl}
is the maximum length of the path any object's centroid traveled during simulation:
\begin{equation}\label{eq:max_pl}
	c_{l} = \max_{\phi \in \Phi}{\displaystyle\sum_{i=1}^{n}{\lVert \mathbf{p}_{i}(\phi) - \mathbf{p}_{i-1}(\phi)\rVert}}
\end{equation}
where $\mathbf{p}_{i}(\phi) \in \mathbb{R}^{3}$ is a position of passive object $\phi$ during the simulation step (= between time steps $t$ and $t+1$) and $n$ is the number of positions covered during simulation. This takes into account different directions of movement, but not a possible spin the object may have been exposed to.

\subsection{Volume-Based Cost Functions}
The following cost functions are based on the volume of the objects instead of their pose, thus considering complex motion paths as well as changes of movement direction and spin.
\setcounter{paragraph}{0}
\paragraph{Maximum Swept Convex Volume $c_{v}$}\label{cost_evaluation:mscv}

\begin{figure}[tb]
  \centering
  \subfigure[initial configuration]{
    \label{fig:mscv1}
    \includegraphics[width=0.135\textwidth]{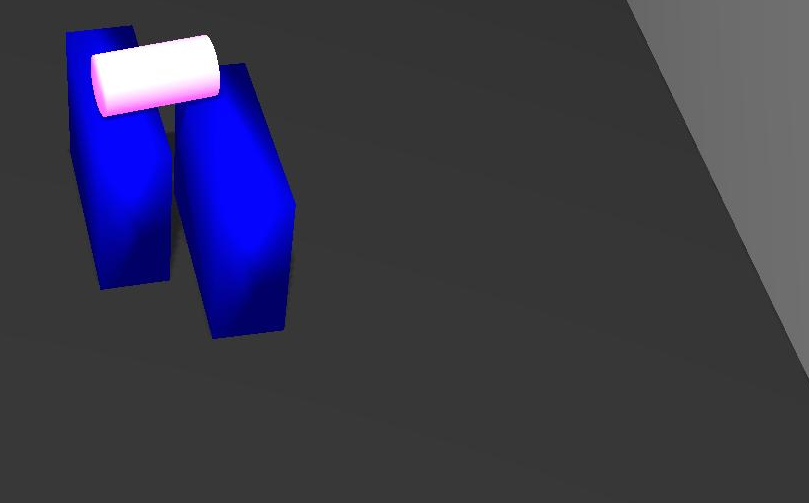} 
  }
  \subfigure[final configuration]{
    \label{fig:mscv2}
    \includegraphics[width=0.135\textwidth]{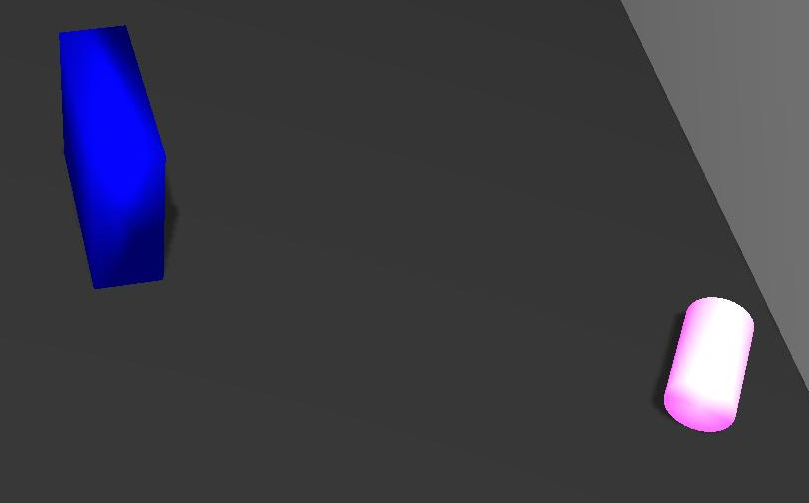} 
  }
  \subfigure[Pose Shift (dashed line), Path Length (dotted line) and Swept Convex Volume (yellow)]{
    \label{fig:mscv3}
    \includegraphics[width=0.295\textwidth]{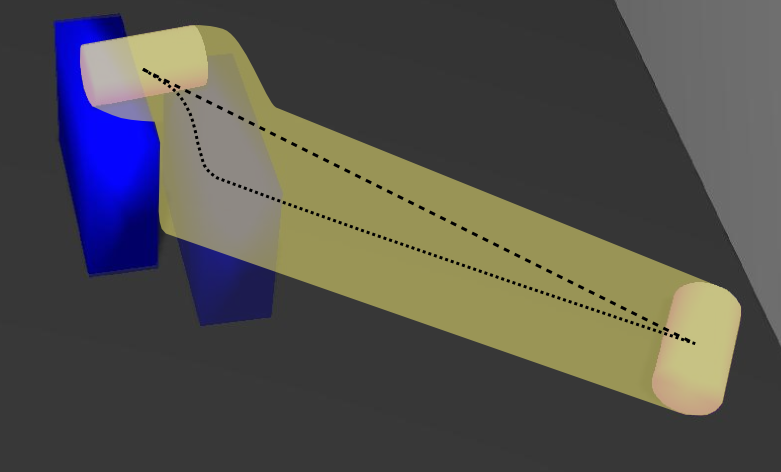} 
  }
  \caption{Visualization of the Swept Convex Volume compared to the Pose Shift and Path Length covered by a passive object during manipulation. When the right box was pulled out by the robot, the cylindrical container fell down and rolled away, causing costs like shown.}
  \label{fig:mscv}
\end{figure}

Swept Volume estimations can be used for collision detection \cite{Baeuml2011} as well as space occupancy calculation for automation and production purposes \cite{VonDziegielewski2012}. For a mathematical formulation see the survey of Abdel-Malek et al.\ \cite{Abdel-Malek2006}.

The object's surface mesh is used as the \emph{generator} which follows a \emph{trajectory} (a set of poses covered while moving during simulation runtime) and creates the \emph{swept volume} (the outer boundary of the object during its motion). This can be considered a voxel-based variant of \cite{Abdel-Malek2006} which is widely used in applications where generator and trajectory are discrete.

In most applications, this volume basically represents a concave hull around all points in space which the object ever touched. However, this takes a lot of effort to compute \cite{VonDziegielewski2012} and, since the concave hull of a number of points is not generally well-defined, may raise ambiguities for different kinds of objects.
As a remedy, we simplified the original Swept Volume approach by replacing the concave with the convex hull, normalized by object volume, which is easy to compute and well-defined, and call the result the \emph{Swept Convex Volume}. Figure~\ref{fig:mscv} shows a visualization of the Swept Convex Volume of a moving object compared to its Pose Shift and Path Length.

Since we relate all objects in a scene, we take the \emph{Maximum Swept Convex Volume} $c_{v}$ which is the maximum over all Swept Convex Volumes of the scene objects. It is determined as
\begin{equation}\label{eq:mscv}
	c_{v} = \max_{\phi \in \Phi}{V(\phi)}
\end{equation}
where $V(\phi)$ is the Swept Convex Volume of the respective object $\phi \in \Phi$ as computed in Algorithm~\ref{alg:scv}. 

\begin{algorithm}[htb]
\centering
\caption{Swept Convex Volume calculation}
\label{alg:scv}
\begin{algorithmic}[1]
    \State \textbf{input}: object mesh $\mathcal{M}(\phi)$, object poses $\mathbf{p}_{0..n}(\phi)$ covered during simulation
    \State create point cloud $\mathcal{C}(\phi)$ from $\mathcal{M}(\phi)$ at $\mathbf{p}_{0}(\phi)$
    \ForAll {$\mathbf{p}_{i}(\phi)$}
        \State $\mathcal{C}_{i}(\phi) \gets \mathcal{C}(\phi)$ transformed from $\mathbf{p}_{0}(\phi)$ to $\mathbf{p}_{i}(\phi)$\label{alg:scv:weight}
        \State $\mathcal{C}(\phi) \gets \mathcal{C}(\phi) \cup \mathcal{C}_{i}(\phi)$
    \EndFor
    \State $H(\phi) \gets \mathrm{convhull}(\mathcal{C}(\phi))$
    \State $H_{0}(\phi) \gets \mathrm{convhull}(\mathcal{C}_{0}(\phi))$
    \State $V(\phi) \gets \mathrm{volume}(H(\phi))~/~\mathrm{volume}(H_{0}(\phi))$
    \State \textbf{output}: Swept Convex Volume $V(\phi)$
\end{algorithmic}
\end{algorithm}

\paragraph{Maximum Weighted Swept Convex Volume $c_{w}$}\label{cost_evaluation:wvol}
Since the desired usage of our method puts special focus on fragile and vulnerable goods, we propose an additional cost function based on $c_{v}$, but with additional weights for each of the 6-D axes.
These weights can be adapted to the domain, e.g. to punish vertically dropping objects like when they fall off a shelf or out of a container. Additionally, one could think of applications like objects running on a conveyor where a lateral translation could push the goods off the belt, thus stopping them to be conveyed. Rotational weights may be employed in a domain dealing with objects which spill when they are turned, for example liquid containers or open underwater objects like amphoras and treasure chests.

As for the implementation of such weights, in Algorithm~\ref{alg:scv} we need to replace $\mathbf{p}_{i}(\phi)$ in Line \ref{alg:scv:weight} with the weighted $\mathbf{p}^{w}_{i}(\phi)$:
\begin{equation}
\mathbf{p}^{w}_{i}(\phi) = \mathbf{p}_{0}(\phi) + \mathrm{diag}(\mathbf{w})\cdot(\mathbf{p}_{i}(\phi)-\mathbf{p}_{0}(\phi))
\end{equation}
where $\mathbf{w}=\begin{bmatrix}w_{x}&w_{y}&w_{z}&w_{\varphi}&w_{\theta}&w_{\psi}\end{bmatrix}^{\intercal}$
are domain-dependent weights for each of the components of the 6-D object pose.

\section{Manipulation Outcome Prediction}\label{outcome_prediction}
Scene Dynamics Anticipation as a validation method for planned actions can be used in runtime-restricted scenarios and scenarios where another planner has already determined which objects to manipulate in which order. One example for this is the planner presented by Mojtahedzadeh et al.~\cite{Mojtahedzadeh2015} which uses static equilibrium calculations to find objects that support each other, hence it will eventually select objects first which do not support any other object.

Implicitly, our approach will usually determine similar manipulation sequences, but in addition considers dynamic events occuring during the manipulation for which Mojtahedzadeh et al.'s approach is not equipped.
Consequently, using our method for validating plans on dynamic scenarios can be seen as an addition and enrichment to another planner.

Algorithm~\ref{alg:outcome_prediction} shows how we perform Outcome Prediction using Scene Dynamics Anticipation with the definitions and cost functions given in the previous sections.

\begin{algorithm}[htb]
\centering
\caption{Manipulation Outcome Prediction}
\begin{algorithmic}[1]
        \State \textbf{input}: scene objects $\mathcal{O}$, active object $\alpha \in \mathcal{O}$
        \State spawn all objects $\mathcal{O}$ in simulation
        \State plan approach $T_{1}^{\alpha}$ and extract $T_{2}^{\alpha}$ trajectory
        \State move simulated robot on $T_{1}^{\alpha}$
        \State grasp active object $\alpha$
        \State move simulated robot on $T_{2}^{\alpha}$
        \State release active object $\alpha$
        \State $c \gets$ manipulation costs ($\rightarrow$ Section~\ref{cost_evaluation})
        \State \textbf{output}: manipulation costs $c$
\end{algorithmic}\label{alg:outcome_prediction}
\end{algorithm}

One drawback of using Scene Dynamics Anticipation as a validation method, however, is the fact that it outputs the manipulation costs for the given scene configuration, but classification between positively and negatively validated actions has to be performed manually via a threshold.

This is a common problem for Outcome Prediction methods in general, also for Rockel et al.'s approach~\cite{Rockel2015} where thresholds are defined manually for which an object is considered to be toppling. Another example is \cite{Pastor2011} where the authors perform statistical tests on whether a planned motion lies within a learned envelope. Whenever such an approach is deployed on a new scenario, these parameters have to be set accordingly.

We empirically determined the manipulation cost threshold based on satisfactory operation of our method. For the logistics scenario, for instance, where the container is low above ground and provides lots of operation space, a \mbox{threshold} of $c_{w}=2.0$ gives satisfactory results. The implicit lower limit for this value is $c_{w}=1.0$ which equals the volume of the object. Since jitter can always occur in simulation and small pushes usually do not cause any damage, setting this value too close to $1.0$ will prevent the algorithm from positively evaluating any configuration. On the other hand, too loose settings should be avoided.

\section{Manipulation Sequence Planning}\label{sequence_planning}
In addition to Manipulation Outcome Prediction, we now want to introduce a way to include Scene Dynamics Anticipation into planning a sequence of manipulation actions.

Instead of serving as a validation tool only, our method is able to plan a manipulation sequence which considers side-effects of moving objects. In addition, obstacle-avoiding motion planning may fail in certain scenarios. But since we explicitly allow for moving passive objects during manipulation, though trying to minimize it, our high-level planner may still succeed in finding a viable manipulation sequence.

In our previous work \cite{Stoyanov2016}, we have sketched the difficulties and pitfalls of motion planning in heavily confined spaces, i.e.\ when all passive objects are considered as obstacles. This may even lead to a dead end in manipulating any remaining object if the constraints are too tight. Manipulation Sequence Planning using Scene Dynamics Anticipation, on the other hand, allows for passive objects to move and takes their motion into account as a feature, thus getting stuck in such a case is less likely.

\subsection{Planning Procedure}
\begin{figure}[tb]
\centering
\newlength\treeheight
\setlength{\treeheight}{\textheight}
\begin{scaletikzpicturetowidth}{0.45\textwidth}
\begin{tikzpicture}
  [
	scale=\tikzscale,
	transform shape,
    level distance          = 12em,
	level 1/.style={sibling distance=\textwidth/3},
	level 2/.style={sibling distance=\textwidth/2},
	level 3/.style={sibling distance=\textwidth/4},
    edge from parent/.style = {draw, -latex},
    every node/.style       = {font=\Huge,align=center},
    sloped
  ]
  \node[top color=olive!100, bottom color=olive!100] {{\includegraphics[width=.2\textwidth]{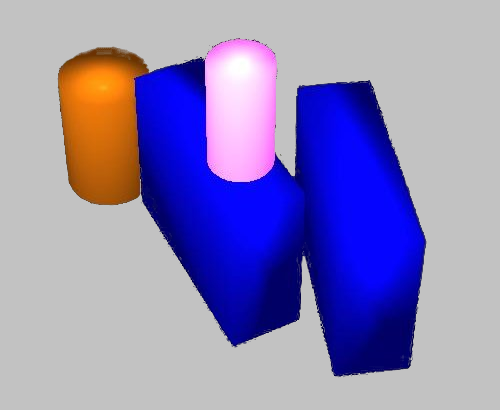}\includegraphics[width=.2\textwidth]{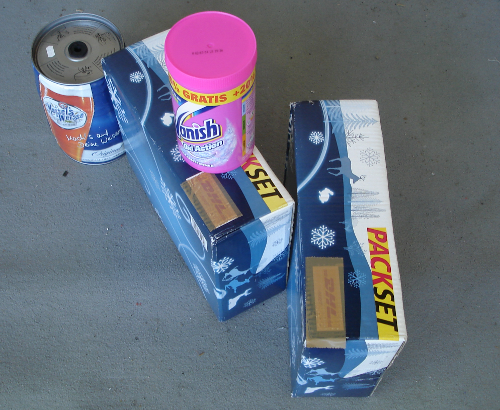}}}
    child { }
    child { }
    child { node[top color=olive!100, bottom color=olive!100] {{\includegraphics[width=.2\textwidth]{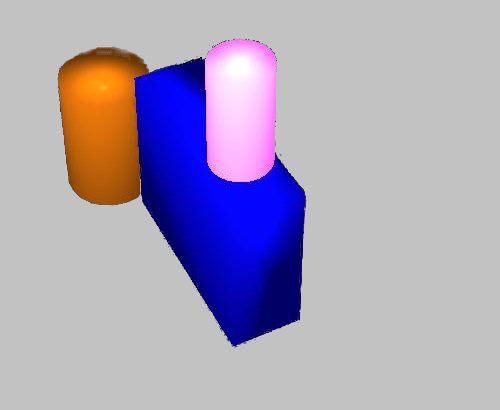}}}
    child { node[top color=olive!100, bottom color=olive!100] {{\includegraphics[width=.2\textwidth]{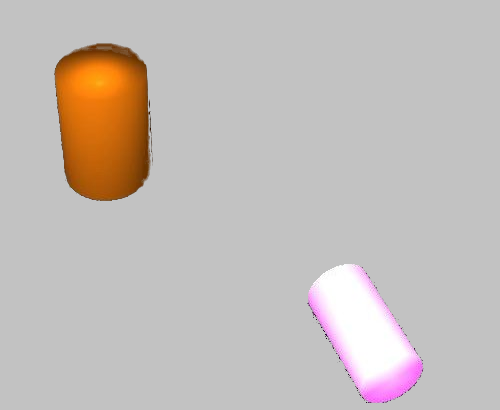}}} 
		child { node[top color=olive!100, bottom color=olive!100] {{\includegraphics[width=.2\textwidth]{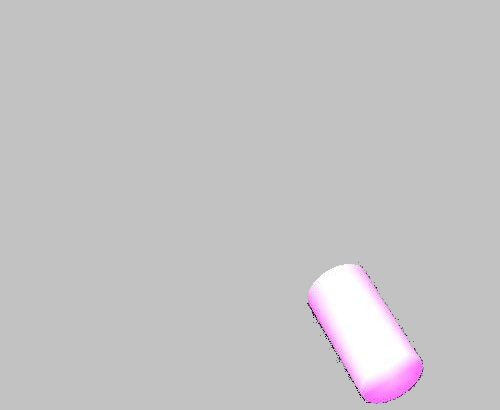}}} }
		child { node[top color=olive!100, bottom color=olive!100] {{\includegraphics[width=.2\textwidth]{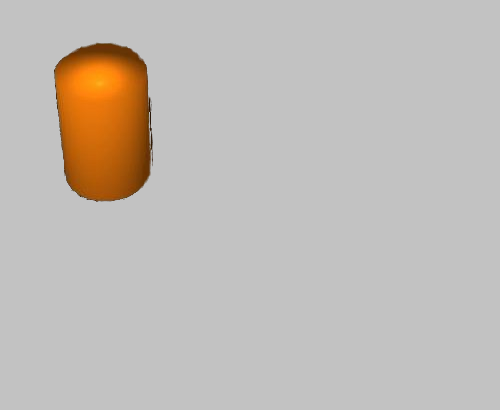}}} } }
    child { node[top color=olive!100, bottom color=olive!100] {{\includegraphics[width=.2\textwidth]{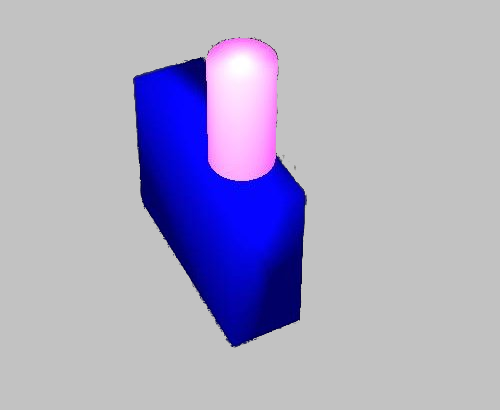}}} 
		child { node[top color=olive!100, bottom color=olive!100] {{\includegraphics[width=.2\textwidth]{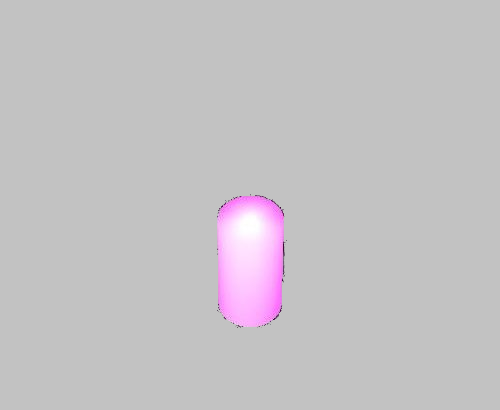}}} }
		child { node[top color=olive!100, bottom color=olive!100] {{\includegraphics[width=.2\textwidth]{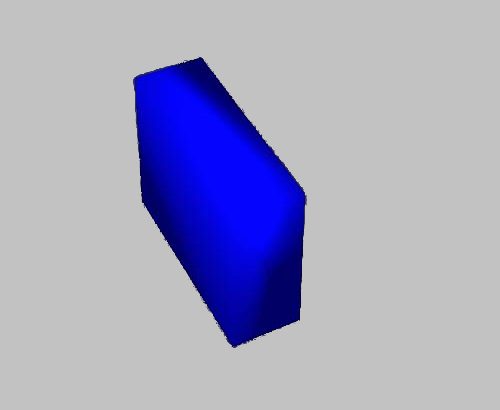}}} } }
    child { node[top color=olive!100, bottom color=olive!100] {{\includegraphics[width=.2\textwidth]{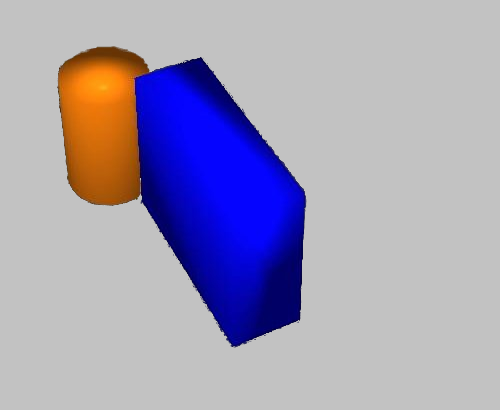}}} 
		child { node[top color=olive!100, bottom color=olive!100] {{\includegraphics[width=.2\textwidth]{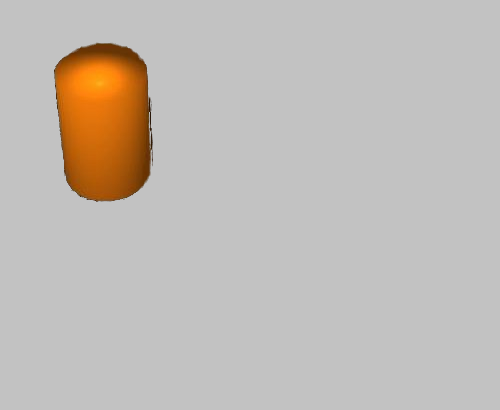}}} }
		child { node[top color=olive!100, bottom color=olive!100] {{\includegraphics[width=.2\textwidth]{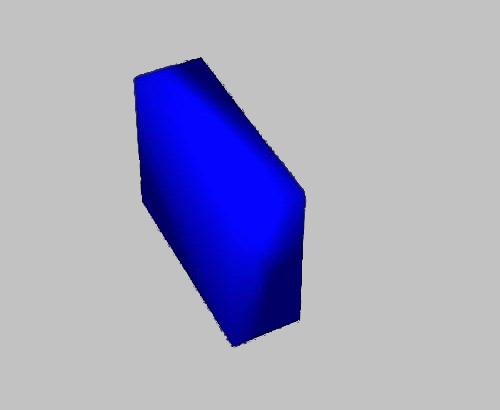}}} } } }
    child { }
;
	\node at (-13,0) {Level 4\\ (1 node)};
	\node at (-13,-4.25) {Level 3\\ (4 nodes)};
	\node at (-13,-8.5) {Level 2\\ (12 nodes)};
	\node at (-13,-12.75) {Level 1\\ (24 nodes)};
\end{tikzpicture}
\end{scaletikzpicturetowidth}
	\caption{Example search tree (other branches cropped for better visibility) -- each node shows its initial configuration, i.e.\ before manipulation}
	\label{fig:tree_example}
\end{figure}

Our method uses a \emph{search tree} built from all possible objects configurations which can occur while manipulating all objects in the scene, see the example in Figure~\ref{fig:tree_example}. The initial configuration is shown as the tree root, on the top, while further down the tree one object has been unloaded in each node. The images show the respective initial configuration for these nodes, before performing the manipulation step.

While traversing the tree further down, unintended motion and high costs are caused for the pink container which falls off the blue box when this one is manipulated.
The leaves merely contain one remaining object which can directly be manipulated disregarding the scene physics since no passive objects are left.

\begin{algorithm}[tb]
\centering
\caption{Search Tree Generation for Sequence Planning}
\begin{algorithmic}[1]
  \State initialize search tree $\mathcal{S}_{0} \gets \emptyset$
  \State initialize set of objects $\mathcal{O}_{0}$ with current scene
	\Procedure{createNode}{$\mathcal{S}_{i}$, $\mathcal{O}_{i}$}
	  \If{$|\mathcal{O}_{i}|\le 1$, i.e.\ this is a leaf node}
		\State\Return{}
	  \Else
		\State create new tree node $N_{i}$ containing object set $\mathcal{O}_{i}$ 
		\State $\mathcal{S}_{i+1} \gets \mathcal{S}_{i} \cup N_{i}$
		\State determine new active object $\alpha_{i} \in \mathcal{O}_{i}$
		\State$\mathcal{O}_{i+1} \gets \mathcal{O}_{i} \setminus \alpha_{i}$
		\State\Return{\Call{createNode}{$\mathcal{S}_{i+1}$, $\mathcal{O}_{i+1}$}}
	  \EndIf
	\EndProcedure
 \State \textbf{output}: filled search tree $\mathcal{S}$
\end{algorithmic}\label{alg:planning_tree_generation}
\end{algorithm}

Algorithm~\ref{alg:planning_tree_generation} shows how to generate the search tree for a scene given the initial object configuration. After the tree has been filled, traversing it works like shown in Algorithm~\ref{alg:planning_anticipation}, using a depth-first search-like technique and exploiting our Manipulation Outcome Prediction method (Algorithm~\ref{alg:outcome_prediction}).

As for the cost functions used for calculating the node costs, all those introduced in Section~\ref{cost_evaluation} can in principle be used interchangeably here. Since one of the main topics of this paper is damage avoidance, however, we prefer the Maximum Weighted Swept Convex Volume $c_{w}$ because it takes into account a) complex motion paths and b) change of movement direction and spin. Both of these factors are a crucial ingredient to avoid objects toppling or dropping under all circumstances.

In order to use $A^{*}$ or a similar search algorithm for our problem we would need to design an admissible heuristic. 
However, with the cost function used in our method it is not possible to define such a heuristic because the distance (in costs) to the target configuration cannot be determined.

\begin{algorithm}[tb]
\centering
\caption{Manipulation Sequence Planning}
\label{alg:planning}
\begin{algorithmic}[1]
    \State \textbf{input}: scene objects $\mathcal{O}$, search tree $\mathcal{S}$ generated from $\mathcal{O}$ ($\rightarrow$ Algorithm~\ref{alg:planning_tree_generation})
    \ForAll {nodes $N_{i} \in \mathcal{S}$}
		\State perform Outcome Prediction ($\rightarrow$ Algorithm~\ref{alg:outcome_prediction})
	\EndFor
        \ForAll {leaf nodes $N_{j}^{\mathrm{leaf}} \in \mathcal{S}$}
		\State $c_{j} \gets$ summed-up costs of $N_{j}^{\mathrm{leaf}}$'s parents
		\State $c_{\mathrm{min}} \gets \mathrm{min}(c_{\mathrm{min}}, c_{j})$
	\EndFor
    \State \textbf{output}: node $N_{j}$ with lowest manipulation costs $c_{\mathrm{min}}$
\end{algorithmic}\label{alg:planning_anticipation}
\end{algorithm}

\begin{figure*}[tb]
  \centering
  \subfigure[Scene 1]{
    \label{fig:scene1}
	\includegraphics[height=0.13\textwidth]{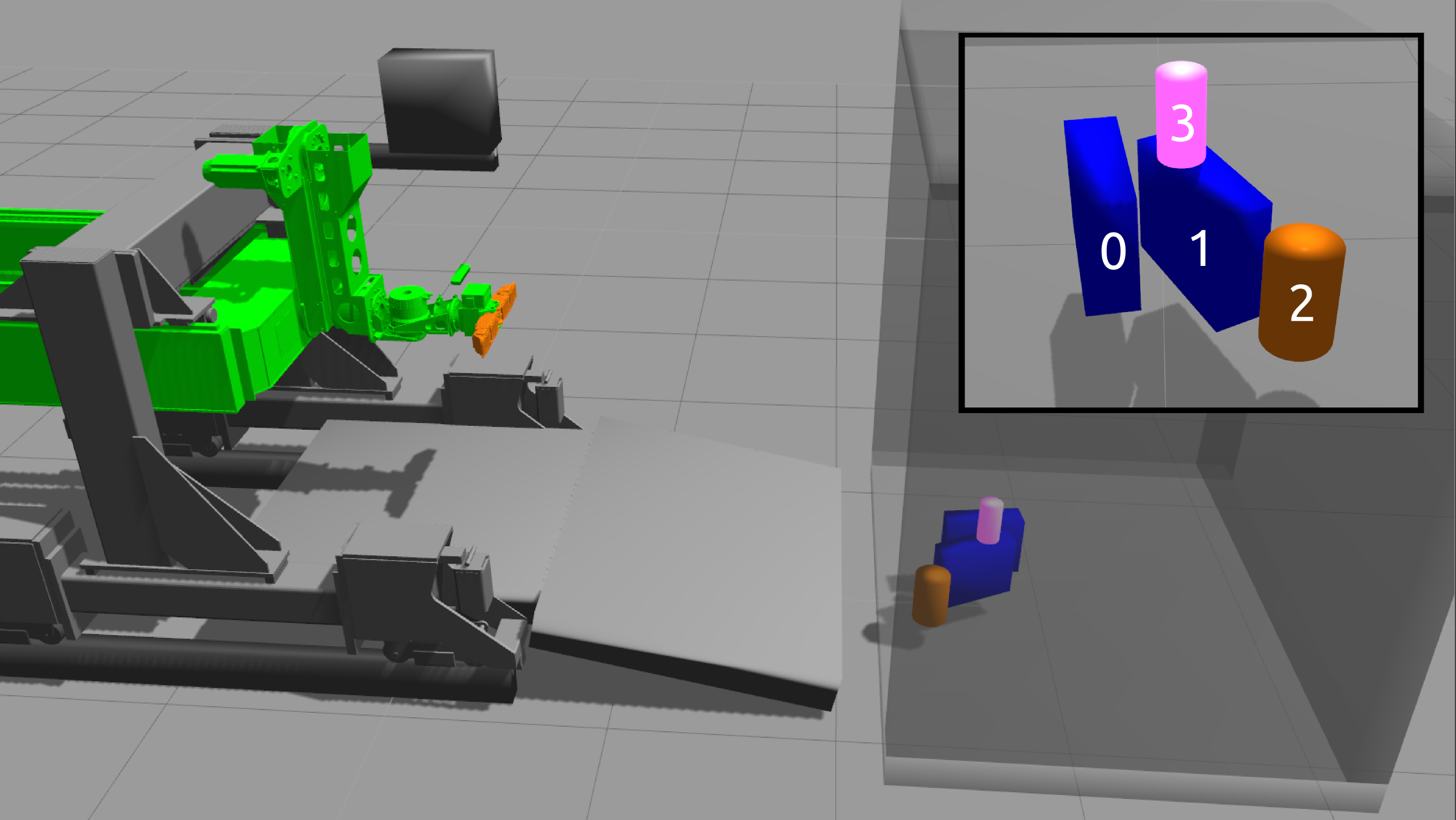} 
  }
  \subfigure[Scene 2]{
    \label{fig:scene2}
    \includegraphics[height=0.13\textwidth]{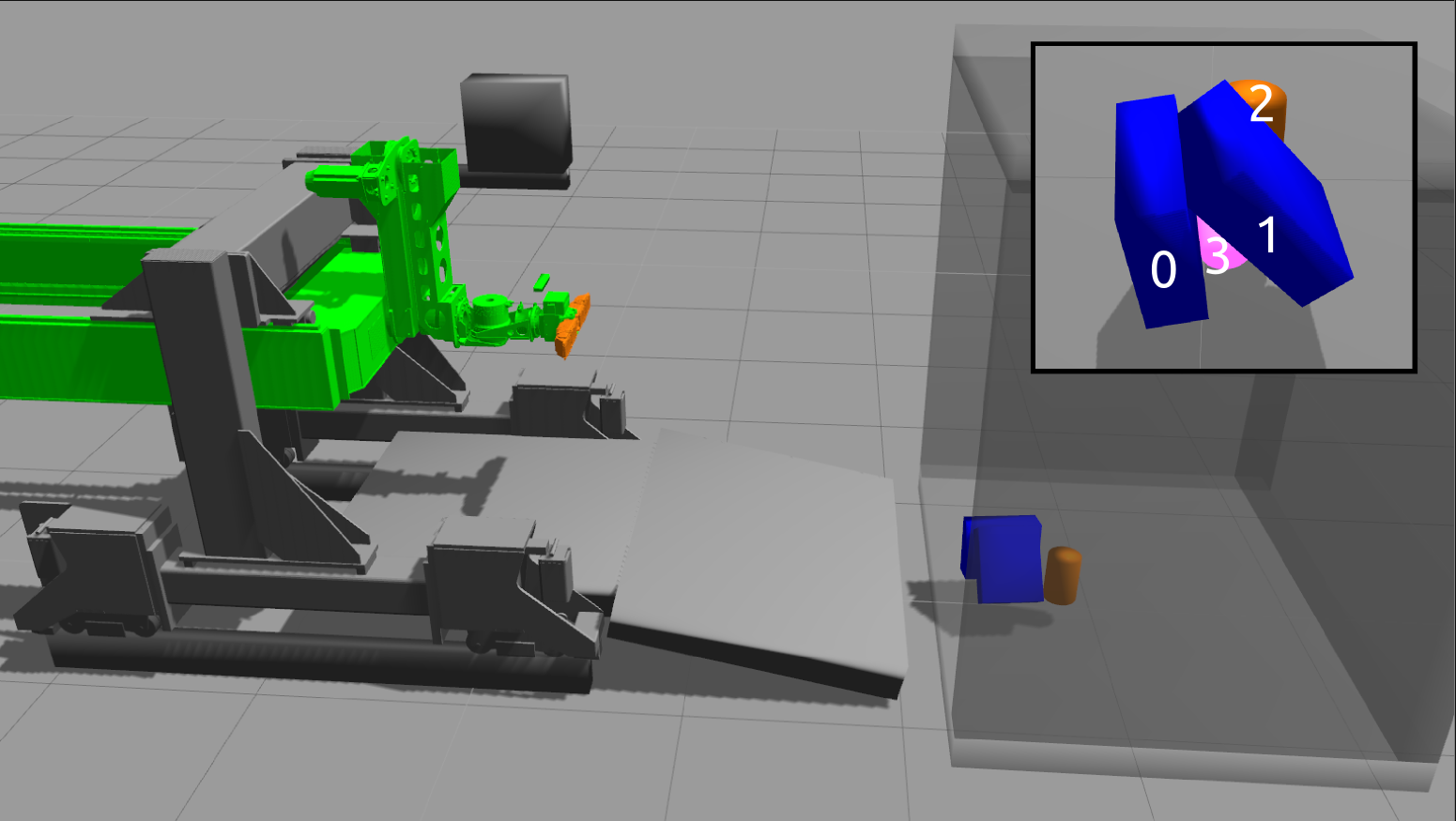} 
  }
  \subfigure[Scene 3]{
    \label{fig:scene3}
    \includegraphics[height=0.13\textwidth]{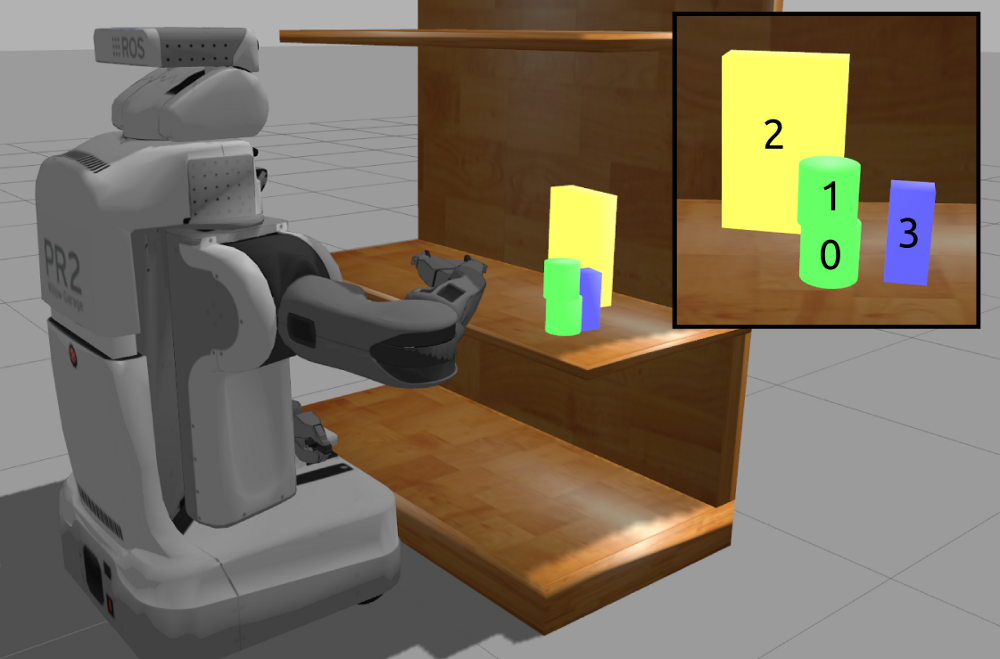} 
  }
  \subfigure[Scene 4]{
    \label{fig:scene4}
    \includegraphics[height=0.13\textwidth]{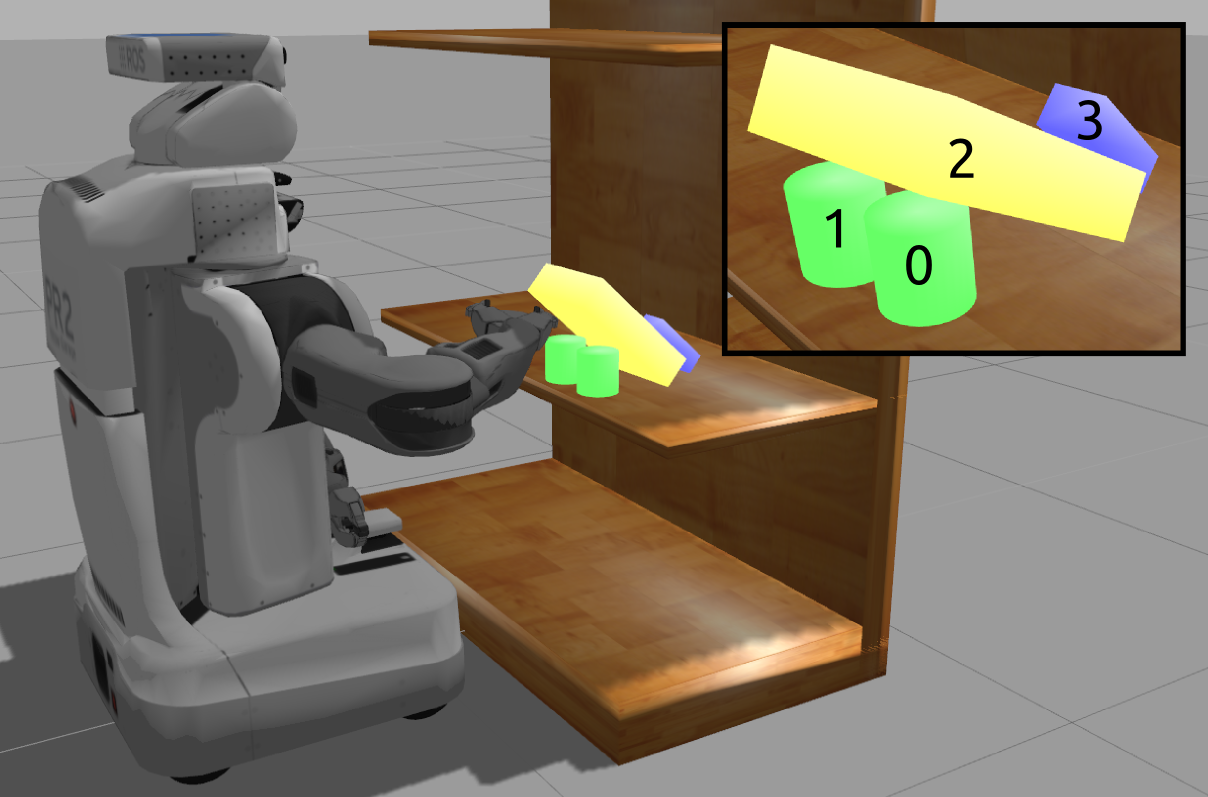} 
  }
  \caption{Experimental scenes used for evaluating our method}
  \label{fig:experiment_scenes}
\end{figure*}

\subsection{Efficiency Considerations}
In contrast to $A^{*}$ or other common search methods like discretized Rapidly-Exploring Random Trees (RRT) or Rapidly-Exploring Random Leafy Trees (RRLT) \cite{Morgan2004}, our scenario causes more of an efficiency issue in estimating the node costs than in the search itself. Traversing the tree taken by itself is rather trivial and fast because the number of nodes which are handled in our search trees can be processed rapidly on modern machines. Thus, the efficient search problem is overshaded by a heuristic cost estimation problem during the processing of each individual node which is prominent and crucial to solve in minimal time.

The \emph{worst-case} tree size for a number of objects $n$, constructed from all possible permutations of the object set like in Algorithm~\ref{alg:planning_tree_generation}, is
\begin{equation}
	\sum_{i=1}^{n} \frac{n!}{i!} = n!+\frac{n!}{2}+...+\frac{n!}{(n-1)!}+1
\end{equation}
where $\frac{n!}{i!}$ is the number of objects for the respective tree level $i$, $i=1$ representing the bottom level (i.e.\ leaves) and $i=n$ the top level (i.e.\ root); see the example in Figure~\ref{fig:tree_example}.

This number grows fast with a rising number of objects, taking along the cost estimation problem with the same speed. A possible remedy for this is to partition the scene into piles of 4-5 objects and plan a manipulation sequence for each of these piles individually. However, how to optimize these partitions still remains an open research question.

In addition to not using more than 4-5 objects at a time, we cut off as many branches of the search tree as possible in which ultimately there is no possibility to present the optimal solution. In the following, we will present several ideas to reduce the tree size as far as possible. In Section~\ref{experiments} we show that the tree size in a typical scenario can be reduced to as low as 48.7\% using these measures.

\subsubsection{Implicit Search Tree Pruning}
As we will show by an empiric consideration in the Results section, in every scene there is a number of configurations which implicitly cannot be simulated. Usual reasons for this are that approaching an object may push a passive object which, in turn, moves the active object away by a significant distance. In this case, the active object will end up unreachable for the planned manipulation action.
Additionally, sometimes no feasible grasping configuration can be found without the robot colliding with the environment (container, shelf, etc.). Since motion planning is out of scope of our work, we have to skip simulating the respective configuration in this case and impose infinite costs.

\subsubsection{Explicit Search Tree Pruning}
In addition to the implicit reasons given above, the used depth-first search allows for cropping tree branches: After simulating a configuration, if the accumulated costs in the current branch exceed the total costs of any already computed goal sequence (i.e.\ a tree leaf), we stop exploring the current branch.

\subsubsection{Re-Use of Similar Configurations}
Another way to reduce computational load, in addition to pruning tree branches, is to re-use similar configurations which occured before somewhere else in the tree. Before running the simulation, we compare the current configuration to each one which was simulated before, anywhere in the tree, for similarity of objects and their poses. If there is a similar configuration, we re-use this node and the whole child tree of the node (if any) without having to simulate any of them.

Generally speaking, apart from the mentioned measures to reduce computational complexity, the runtimes of our method strongly depend on the used simulator. Moreover, regarding the total runtime of one sequence planning run, motion planning and execution take the biggest part of time which is very robot and application-specific.

Fortunately, our method is well-suited for parallelization because all nodes on the same tree level can be simulated in parallel since their configurations do not depend on each other. As the number of objects in the scene and, with it, the number of tree levels rises, the limit for the speedup achieved by parallelization is the number of used threads / CPU cores.

\subsection{Open-Loop vs. Closed-Loop Planning}
The method we propose basically allows for open-loop planning, where the manipulation sequence is planned once and then executed without further anticipation of scene dynamics, or closed-loop, where the whole process is repeated from the beginning after manipulation of any object.
Re-planning the entire remaining manipulation sequence after each perception/manipulation cycle in a closed loop has the advantage of covering object movement which stems from to inaccurate physics simulation or other, external disturbances in the scene. However, time constraints may not allow for running Manipulation Sequence Planning after each manipulation cycle, so using the results only as an initial plan is a viable alternative.

\section{Applications and Experimental Results}\label{experiments}
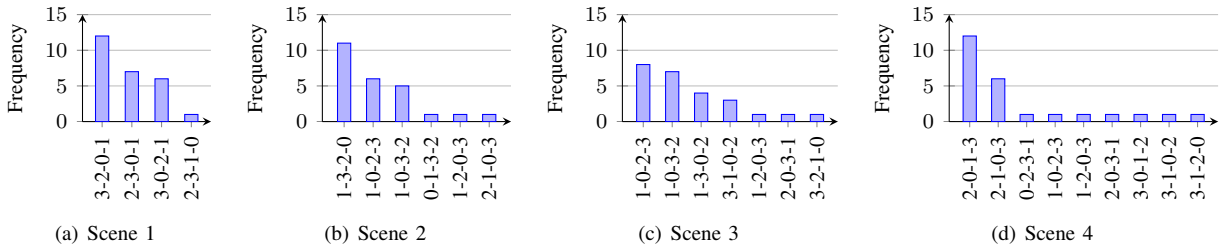
\begin{figure*}[btp]
  \centering
\subfigure[Scene 1]{
\begin{tikzpicture}[font=\footnotesize]
\begin{axis}[
      ybar,
      bar width=5pt,
	  ymajorgrids=true,
      ylabel={Frequency},
      ymin=0,
      ymax=15,
      xtick=data,
	  xticklabel style={rotate=90},
      axis x line=bottom,
      axis y line=left,
	  ylabel near ticks,
      enlarge x limits=0.22,
      symbolic x coords={3-2-0-1,2-3-0-1,3-0-2-1,2-3-1-0},
	  width=0.185\textwidth,
	  height=0.125\textheight
    ]
	\addplot+[]file{histogram_scene1_first.csv};
    \end{axis}
\end{tikzpicture}
}
\subfigure[Scene 2]{
\begin{tikzpicture}[font=\footnotesize]
\begin{axis}[
      ybar,
      bar width=5pt,
	  ymajorgrids=true,
      ylabel={Frequency},
      ymin=0,
      ymax=15,
      xtick=data,
	  xticklabel style={rotate=90},
      axis x line=bottom,
      axis y line=left,
	  ylabel near ticks,
      enlarge x limits=0.15,
      symbolic x coords={1-3-2-0,1-0-2-3,1-0-3-2,0-1-3-2,1-2-0-3,2-1-0-3},
	  width=0.23\textwidth,
	  height=0.125\textheight
    ]
	\addplot+[]file{histogram_scene2a_first.csv};
    \end{axis}
\end{tikzpicture}
}
\subfigure[Scene 3]{
\begin{tikzpicture}[font=\footnotesize]
\begin{axis}[
      ybar,
      bar width=5pt,
	  ymajorgrids=true,
      ylabel={Frequency},
      ymin=0,
      ymax=15,
      xtick=data,
	  xticklabel style={rotate=90},
      axis x line=bottom,
      axis y line=left,
	  ylabel near ticks,
      enlarge x limits=0.12,
      symbolic x coords={1-0-2-3,1-0-3-2,1-3-0-2,3-1-0-2,1-2-0-3,2-0-3-1,3-2-1-0},
	  width=0.25\textwidth,
	  height=0.125\textheight
    ]
	\addplot+[]file{histogram_scene3_first.csv};
    \end{axis}
\end{tikzpicture}
}
\subfigure[Scene 4]{
\begin{tikzpicture}[font=\footnotesize]
\begin{axis}[
      ybar,
      bar width=5pt,
	  ymajorgrids=true,
      ylabel={Frequency},
      ymin=0,
      ymax=15,
      xtick=data,
	  xticklabel style={rotate=90},
      axis x line=bottom,
      axis y line=left,
	  ylabel near ticks,
      enlarge x limits=0.09,
      symbolic x coords={2-0-1-3,2-1-0-3,0-2-3-1,1-0-2-3,1-2-0-3,2-0-3-1,3-0-1-2,3-1-0-2,3-1-2-0},
	  width=0.29\textwidth,
	  height=0.125\textheight
    ]
	\addplot+[]file{histogram_scene4_first.csv};
    \end{axis}
\end{tikzpicture}
}
\caption{Frequencies of first-ranked manipulation sequences selected by our method}
\label{fig:histograms}
\end{figure*}

In this section, we want to evaluate our method applied on two of the described practical scenarios. Typical scenes for both are shown in Figure~\ref{fig:scenarios}.

As for the first scenario, logistics is a field where many different goods have to be handled, some of which are heavy, bulky, fragile or otherwise damage-prone. Our previous work on unloading shipping containers \cite{Stoyanov2016} provides an ideal application in this respect, so we have modeled the robot, container and objects from this scenario as our first example.

Secondly, domestic and retail robotics provide another playground for manipulation strategy planning, thus the supermarket scenario of \cite{Winkler2016} will serve as the second example.

\subsection{Prerequisites}
For evaluating our method, we take several modalities as granted and thus abstract from them because they fall out of the scope of this work.

First of all, since the perception and grasping accuracy of our previous work have been validated extensively already in \cite{Stoyanov2016}, \cite{Vaskevicius2014}, \cite{Vaskevicius2012} and \cite{Mueller2014}, we are not going into detail on these measures anymore.

Abstracting from details in the object models is necessary as well because we would like to show the general concept of our method in this work. Although it is desirable to obtain simulation models as close as possible to the real objects, this constitutes an own branch of research. Because of this, we use a simplified representation of the rigid objects from our real scenarios (see Figure~\ref{fig:scenarios}) like shown in Figure~\ref{fig:experiment_scenes}.
For the same reason, we do not employ complex grasping techniques here, but instead generate simple grasping configurations around the principal axes.

Although we present a general approach which is not limited to scenarios where damage-awareness is the main focus, we want to emphasize the importance of this factor in our applications. Thus, we use the Maximum Weighted Swept Convex Volume $c_{w}$ as a cost function which explicitly penalizes covered space. The weights $\mathbf{w}$ for $c_{w}$ we set on
\begin{equation*}
\mathbf{w} = \begin{bmatrix}1.0&1.0&2.0&1.0&1.0&1.0\end{bmatrix}^{\intercal}
\end{equation*}
which puts increased importance on damage-prone vertical motion. This is desired especially in scenarios like a supermarket where objects would break if dropped from a shelf.

\subsection{Experimental Scenes}

We want to show how our method works by running it many times on different scenes and comparing the final manipulation sequences. Additionally, we show what happens to the search tree which can be heavily pruned during traversal.

\paragraph{Logistics}
Scenes 1 and 2 in Figure~\ref{fig:experiment_scenes} show typical scenarios encountered in RobLog \cite{Stoyanov2016} where Scene 2 is the more challenging one due to the goods supporting each other. Hence, once any of the objects is manipulated it is likely that one of the others will move as well.

\paragraph{Supermarket}
The tall shelf, along with the cans which may roll away if dropped, is a hostile area for any damage-prone product and thus a good example for our method. The tipped-over items in Scene 4 occur frequently in a supermarket where customers leave the shelf like this, products coming to rest partly on top of each other.

\subsection{Manipulation Sequence Planning Results}
We ran our method 25 times each on the two scenes per application, resulting in a total of 100 runs. Figure~\ref{fig:histograms} shows a histogram of how often a particular manipulation sequence was selected for a particular scene of Figure~\ref{fig:experiment_scenes}.

Numerical results are shown in Table~\ref{table:statistics} with their means ($\mu$) and standard deviations ($\sigma$) for the following criteria:
\begin{itemize}
	\item \emph{mean first-ranked costs per node}: mean over 25 runs of the mean costs per object imposed on the manipulation sequence selected by Algorithm~\ref{alg:planning_anticipation}
	\item \emph{mean second-ranked costs per node}: mean over 25 runs of the mean costs per object imposed on the second-best manipulation sequence
	\item \emph{pruned tree nodes}: percentage of nodes pruned from the tree, resulting in a similar reduction in runtime, and composed of the following sub-criteria:
	\begin{itemize}
		\item \emph{known subtree}: a node shares its object configuration and poses with a previously processed node, thus the costs were copied without re-simulating
		\item \emph{costs exceed existing sequence}: a solution is existing already which has lower costs than the costs accumulated so far in this branch
		\item \emph{active object moved}: the active object was pushed away during approach, ending up unreachable
		\item \emph{object out of workspace}: an object fell out of the workspace (container/shelf), causing maximum damage and ending up unreachable for the robot
		\item \emph{planning failure}: no motion plan was found for the active object, e.g.\ because it was pushed away too far in a parent node
	\end{itemize}
	\item \emph{nodes with significant movement}: percentage of nodes which were not pruned from the tree and reported significant costs higher than a manually defined threshold 
\end{itemize}

\begin{table*}[btp]
\centering
\caption{Numerical Results}
\begin{tabular}{l|r|r|r|r|r|r|r|r||r|r}
\textbf{Scene} & \multicolumn{2}{c|}{\textbf{1}} & \multicolumn{2}{c|}{\textbf{2}} & \multicolumn{2}{c|}{\textbf{3}} & \multicolumn{2}{c||}{\textbf{4}} & \multicolumn{2}{c}{\textbf{Total}} \\
\hline
& $\mu$ & $\sigma$ & $\mu$ & $\sigma$ & $\mu$ & $\sigma$ & $\mu$ & $\sigma$ & $\mu$ & $\sigma$ \\
\hline
mean first-ranked costs per node       &  1.008 &  0.008 &  1.208 &  0.074 &  1.349 &  0.359 &  1.975 &  0.321 &  1.382 &  0.435 \\
mean second-ranked costs per node      &  1.337 &  0.868 &  1.429 &  0.326 &  1.700 &  1.728 &  3.764 &  7.393 &  2.051 &  3.878 \\
\hline
pruned tree nodes                      & 34.8\% &        & 50.4\% &        & 58.3\% &        & 61.9\% &        & 51.3\% &        \\
 - known subtree                       &  7.5\% &  4.8\% &  2.6\% &  2.5\% &  1.2\% &  2.2\% &  0.0\% &  0.0\% &  2.8\% &  4.1\% \\
 - costs exceed existing sequence      & 16.5\% &  8.3\% & 18.3\% &  5.3\% & 19.3\% & 14.7\% & 41.0\% & 18.1\% & 23.8\% & 16.0\% \\
 - active object moved                 &  5.8\% &  5.6\% & 25.2\% & 13.5\% &  4.2\% & 11.6\% &  8.1\% &  9.2\% & 10.9\% & 13.2\% \\
 - object out of workspace             &  0.0\% &  0.0\% &  0.0\% &  0.0\% & 31.9\% & 15.7\% &  7.0\% &  9.6\% &  9.6\% & 15.9\% \\
 - planning failure                    &  5.0\% &  9.6\% &  4.3\% &  8.4\% &  1.7\% &  3.1\% &  5.8\% & 10.9\% &  4.2\% &  8.6\% \\
\hline
\emph{for reference:} non-pruned nodes with  &  &        &        &        &        &        &        &        &        &        \\
significant movement ($c_{w}>2.0$)     & 13.8\% &  3.0\% & 11.9\% &  4.9\% & 14.8\% &  4.3\% & 18.9\% &  5.1\% & 14.9\% &  5.0\% \\
\end{tabular}
\label{table:statistics}
\end{table*}

\subsection{Discussion}
The results in Figure~\ref{fig:histograms} show that, over many runs, of the possible number of sequences only a small selection is considered as best. The fact that not the same sequence is reported for each run, but there is obviously some noise in the selection stems mostly from the used motion planning techniques. These rely on random trees and thus deliver noisy results for similar inputs. This way, different grasping configurations are selected for the same object configuration, resulting in different motions even if the scene looks similar.

Nevertheless, the resulting manipulation sequences are valid and reasonable in a way that they comply with the intuitive sequence a human would use to clear the container/shelf: unloading the objects first which do not physically support others. So, even if there is some distribution of the finally selected manipulation sequence due to reasons that cannot be influenced by our planning approach, our method succeeds in selecting reasonable solutions for all scenes.

The results show that generally a major ratio of pruned nodes results from comparing the current costs to the already planned sequences. Therefore, this is one of the most useful tricks we use to reduce computational load.

In Scene 3, items falling off the shelf are the major reason for tree pruning: 31.9\% of nodes featured an object which moved out of the robot's workspace. The robot used in the logistics scenario has a big workspace in all directions, additionally, the container provides physical limits on three sides, thus this did not happen for Scenes 1 and 2.

The high values for the standard deviations of pruned nodes result from all the children of such a node automatically being cut off. This way, any disturbance propagates down through the tree and may affect many nodes at the same time when it happens in a high-level node.

\section{Conclusion}\label{sec:conclusion}
We presented a method to plan manipulation sequences for a number of objects, depending on their physical behavior. This enables autonomous robots to anticipate an optimal manipulation order to avoid potential damage by taking into account unintended movement of other objects.

Our approach presents merely a first attempt to solve a complex problem. However, since the planning technique is abstract, given the scene was perceived correctly, it is universally usable on any autonomous robot scenario.

We evaluated our approach on two everyday scenarios showing good results for scenes of different physical complexity and on different types of robots. 

\bibliographystyle{IEEEtran}
\bibliography{bibliography}
\end{document}